\documentclass[pdflatex,sn-apa]{sn-jnl}
\usepackage[utf8]{inputenc}
\usepackage{CJKutf8}
\usepackage{multirow}
\usepackage{amsmath,amssymb,amsfonts}
\usepackage{amsthm}
\usepackage{mathrsfs}
\usepackage[title]{appendix}
\usepackage{xcolor}
\usepackage{textcomp}
\usepackage{manyfoot}
\usepackage{booktabs}
\usepackage{algorithm}
\usepackage{algorithmicx}
\usepackage{algpseudocode}
\usepackage{listings}
\usepackage{booktabs}
\usepackage{diagbox}
\usepackage{float}

\definecolor{BasicBlue}{RGB}{0, 0, 255}
\hypersetup{
    colorlinks,
    linkcolor = BasicBlue,
    citecolor = BasicBlue,
    urlcolor = BasicBlue }

\makeatletter\def\Hy@Warning#1{}\makeatother 
\usepackage{microtype} 
\usepackage{anyfontsize} 

\begin{document}
\title[Comparison of Scoring Rationales]{Comparison of Scoring Rationales Between Large Language Models and Human Raters}

\author*[1]{\fnm{Haowei} \sur{Hua}}\email{jack.hua@princeton.edu}

\author[2]{\fnm{Hong} \sur{Jiao}}\email{hjiao@umd.edu}

\author[3]{\fnm{Dan} \sur{Song}}\email{dan-song@uiowa.edu}

\affil*[1]{\orgname{Princeton University}, \orgaddress{\street{1 Nassau Hall}, \city{Princeton}, \postcode{08544}, \state{NJ}, \country{USA}}}

\affil[2]{\orgdiv{Department of Human Development and Quantitative Methodology}, \orgname{University of Maryland}, \orgaddress{\street{3942 Campus Dr}, \city{College Park}, \postcode{20742}, \state{MD}, \country{USA}}}

\affil[3]{\orgdiv{Department of Psychological and Quantitative Foundations}, \orgname{University of Iowa}, \orgaddress{\street{361 Lindquist Center}, \city{Iowa City}, \postcode{52242}, \state{IA}, \country{USA}}}

\abstract{Advances in automated scoring are closely aligned with advances in machine-learning and natural-language-processing techniques. With recent progress in large language models (LLMs), the use of ChatGPT, Gemini, Claude, and other generative-AI chatbots for automated scoring has been explored. Given their strong reasoning capabilities, LLMs can also produce rationales to support the scores they assign. Thus, evaluating the rationales provided by both human and LLM raters can help improve the understanding of the reasoning that each type of rater applies when assigning a score. This study investigates the rationales of human and LLM raters to identify potential causes of scoring inconsistency. Using essays from a large-scale test, the scoring accuracy of GPT-4o, Gemini, and other LLMs is examined based on quadratic weighted kappa and normalized mutual information. Cosine similarity is used to evaluate the similarity of the rationales provided. In addition, clustering patterns in rationales are explored using principal component analysis based on the embeddings of the rationales. The findings of this study provide insights into the accuracy and ``thinking'' of LLMs in automated scoring, helping to improve the understanding of the rationales behind both human scoring and LLM-based automated scoring.}

\keywords{Automated essay scoring, Large language model, Natural language processing, Cosine similarity, PCA}

\maketitle

\section{Introduction}
Automated scoring has the potential to be an efficient, scalable, and consistent alternative to human scoring. Human scoring is labor-intensive and costly, with raters often exhibiting variability due to fatigue, personal biases, or subjective interpretations of rubrics \citep{bejar2011validity}. In contrast, automated scoring can use machine-learning algorithms or structured guidelines to objectively evaluate responses \citep{williamson2006automated}. These advantages have led to growing use of automated scoring in standardized assessments such as the Graduate Record Examination (GRE), Test of English as a Foreign Language (TOEFL), and Advanced Placement (AP) exams~\citep{zhang2013contrasting}; however, despite their potential benefits, little is known about how automated scoring engines ``think,'' including their interpretability and how well they align with human raters' scoring rationales, particularly when evaluating complex, open-ended responses such as essays \citep{shermis2013contrasting}.

The evolution of automated scoring has closely followed developments in machine learning and natural language processing (NLP). Early automated essay scoring (AES) systems, such as Project Essay Grade in the 1960s, relied on hand-crafted linguistic features such as word count and syntax patterns \citep{foltz1999intelligent}. In the 1990s and 2000s, more advanced NLP-based systems such as e-rater and the Intelligent Essay Assessor emerged, which incorporated latent semantic analysis and rule-based scoring techniques \citep{attali2006automated}. While these systems represented improvements in terms of accuracy, they continued to face challenges in replicating human evaluative reasoning, particularly for nuanced aspects of writing such as argument strength and logical coherence. More recent deep-learning models, especially transformer-based architectures such as BERT (Bidirectional Encoder Representations from Transformers) and DeBERTa (Decoding-enhanced BERT with disentangled attention), have improved automated scoring by capturing richer semantic and contextual relationships within text, achieving high agreement rates with human scorers \citep{ludwig2021automated}.

With the rise of large language models (LLMs) such as OpenAI's ChatGPT and Google's Gemini, a new approach to automated scoring is emerging. Unlike traditional AES models, which rely on predefined linguistic features, LLMs can generate scores using deep neural networks trained on massive datasets \citep{wang2024beyond}. One advantage of LLM-based scoring is its ability to provide rationales or explanations to justify a given score, offering greater transparency and interpretability than conventional automated scoring methods. Recent research has suggested that LLMs can not only predict the scores assigned by human raters but also produce qualitative reasoning to support those scores \citep{xiao2025human}; however, further study of score accuracy is needed, particularly when scores can vary across prompts and writing styles \citep{tang2024harnessing}.

\section{Related Work}
\citet{powers2000comparing} examined the consistency between AI and human scoring, the generation of scoring rationales, and the types of errors made by automated systems. Early research showed that AES could achieve correlations with human graders that were comparable to the agreement between two human raters. In their study, the trained artificial scorer demonstrated a strong relationship with human scores, indicating the potential of AES to replicate human ratings. More recent research has examined the effectiveness of LLMs, such as ChatGPT, in scoring essays. For instance, \citet{mizumoto2023exploring} investigated the use of generative-AI-based language models for essay scoring and found that ChatGPT can produce evaluations that closely align with human feedback, particularly in terms of grammatical and surface-level features. Similarly, \citet{organisciak2023beyond} found that LLMs can significantly improve the scoring of divergent-thinking tasks compared to traditional semantic-distance approaches, indicating the potential for broader applications in creative and open-ended writing. In addition, \citet{song2024automated} used open-source LLMs for both-essay scoring and revision tasks, producing promising results in terms of capturing linguistic fluency and coherence.

\citet{li2024calibrating} report that various studies have focused on rationale-based scoring, in which LLMs generate textual justifications for the scores they assign. Their findings suggest that incorporating rationales can improve the interpretability of LLM-assigned scores by providing a clearer view of the model's decision-making process. The inclusion of rationale generation also supports a move toward more explainable scoring frameworks, potentially helping to increase transparency in educational settings.

Several recent studies have reported progress in LLM-based scoring systems. \citet{latif2024fine} fine-tuned ChatGPT models for automatic scoring, and they found good agreement with expert human scores, particularly in structured-response tasks. \citet{katuka2024investigating} examined the application of parameter-efficient fine-tuning methods such as low-rank adaptation (LoRA) and quantized low-rank adaptation (QLoRA) to LLaMA-2 models. They found that even quantized LLMs could generate feedback with strong bilingual evaluation understudy (BLEU) and recall-oriented understudy for gisting evaluation (ROUGE) scores, closely matching human-written comments. \citet{lagakis2024evaai} proposed a multi-agent framework, EvaAI, which uses multiple LLMs to collaboratively assess work produced by students across domains. They demonstrated its scalability for grading in massive open online courses and code-based assignments. Additionally, \citet{pack2024large} evaluated the performance of AESs for English-language learners, showing that when LLM-based systems are properly calibrated, they can produce valid and reliable scores across multiple traits, contributing to fairer assessment practices for diverse student groups; however, these models still face challenges in capturing deeper semantic understanding and higher-order reasoning. This is consistent with the concerns raised by \citet{rodriguez2019language}, who noted that while LLMs perform well on surface-level features, their alignment with rubric-based content traits remains inconsistent.

This study contributes to the growing body of research in this field by systematically analyzing scoring rationales from human and LLM raters to identify potential sources of scoring inconsistency. Using evaluation metrics such as quadratic weighted kappa (QWK) and normalized mutual information (NMI), it examines score consistency between LLMs and human raters. Additionally, cosine similarity is used to assess the alignment of scoring rationales between LLM and human raters. The study specifically investigates different scenarios based on similarities and differences between scores and rationales. Finally, principal component analysis (PCA) is applied to explore clustering patterns of embeddings among different LLMs when their scores align with human-assigned scores, highlighting variations in scoring rationales across LLMs. As automated scoring continues to advance, ensuring both scoring accuracy and rationale consistency between LLM and human raters will be important for improving mutual understanding. This will help humans to better interpret the outputs of LLMs while also enabling LLMs to better comprehend human language. Together, these improvements will support wider adoption of LLMs in automated scoring.

\section{Methodology}
\subsection{Dataset}
This study used a dataset consisting of 30 U.S. college students enrolled in Mandarin Chinese language courses, including 18 third-year and 12 fourth-year students from diverse majors such as business, humanities, STEM, and diplomacy. Data were collected over two consecutive days using free-response writing prompts from the 2021 and 2022 AP Chinese exams developed by the College Board. On the first day, students responded to the 2021 prompts, which included one story-narration (SN) task and one email-response (ER) task. On the second day, they completed the 2022 prompts, which also consisted of one SN task and one ER task. Two certified AP Chinese raters evaluated the writing samples using the College Board's scoring guidelines, assessing task completion, delivery, and language use. Each essay received one holistic and three domain-specific scores, each ranging from 0 to 6 \citep{song2025exploring}.

Subsequently, human rater~1, a certified AP Chinese Language and Culture Exam rater, trained seven LLMs---GPT-3.5, GPT-4.0, GPT-4o, Gemini~1.5, Gemini~2.0, Claude~3.5 Sonnet, and OpenAI~o1---using the few-shot prompting method to grade student ER2 essays (see Appendix~A). Few-shot training examples were sourced either from the College Board website or from samples previously scored by human rater~1. For this study, all rationales provided by the seven LLMs for the 30~ER2 student essays were included to illustrate the analysis methods.

\subsection{Preprocessing}

A comprehensive preprocessing pipeline was applied to ensure the quality of the text data being used for analysis. First, text normalization was performed by converting all text to lowercase, preventing any inconsistencies that might be caused by capitalization variations. Next, non-English characters were removed, filtering out non-ASCII symbols to retain only standard English text. To further refine the data, punctuation marks and special characters were eliminated, reducing noise and improving the accuracy of tokenization. Additionally, stopword removal was carried out using the Natural Language Toolkit stopword list, excluding common function words that do not contribute to semantic meaning. Beyond standard stopword filtering, a predefined set of domain-specific words (e.g., ``example,'' ``overall,'' ``student,'' ``email'') was removed to reduce redundancy and increase the informativeness of the text. The selection of these words was based on the assessment guidelines and the content of the machine-generated rationales. Finally, excess whitespace was eliminated, ensuring that the processed text maintained a clean and standardized format. This preprocessing pipeline sought to improve text quality and prepare the data for subsequent analysis.

\subsection{Text Embedding}

To calculate cosine similarity and perform PCA, word texts must be converted into numerical vectors. Text embeddings are numerical vector representations of textual data that capture meaning, context, and relationships with other texts. These embeddings transform raw text into high-dimensional vectors that machine-learning models can process efficiently. By mapping text into a continuous vector space, embeddings enable various NLP applications such as document-similarity analysis, semantic search, topic modeling, and clustering. Word embeddings use large-scale language corpora to learn meaningful representations of words and phrases based on their contextual usage \citep{reimers2019sentence}.

Among transformer-based approaches, Sentence-BERT (SBERT) has become a widely used model for generating high-quality sentence and paragraph embeddings. Unlike traditional BERT models, which are not directly optimized for similarity tasks, SBERT employs a Siamese network structure that enables efficient pairwise comparisons using cosine similarity. It produces fixed-size vector representations for sentences, preserving their semantic content and contextual nuances. SBERT embeddings perform well in tasks such as semantic textual similarity, paraphrase mining, and clustering, where capturing deep contextual meaning is important. Furthermore, SBERT supports various pretrained backbones (e.g., RoBERTa and DistilBERT), and it can be fine-tuned for domain-specific tasks, making it a versatile tool for modern NLP pipelines \citep{reimers2019sentence}.

\subsection{Evaluation Metrics}
\subsubsection{Quadratic Weighted Kappa}
The QWK score is a statistical measure that can be used to evaluate the agreement between two sets of categorical ratings while accounting for the ordinal nature of the data. It extends Cohen's kappa by introducing a quadratic weighting scheme that penalizes larger discrepancies more heavily. Given two sets of ratings, \(X\) and \(Y\), the QWK score is computed as
\begin{equation}
\mathrm{QWK} = 1 - \frac{\sum_{i,j} O_{ij} W_{ij}}{\sum_{i,j} E_{ij} W_{ij}}
\end{equation}
where \( O_{ij} \) represents the observed agreement matrix, \( E_{ij} \) is the expected agreement matrix under random chance, and \( W_{ij} \) is the quadratic weighting function,
\begin{equation}
W_{ij} = \frac{(i - j)^2}{(k - 1)^2}
\end{equation}
where \(i\) and \(j\) are rating categories and \(k\) is the total number of categories. This weighting scheme ensures that larger rating mismatches are penalized more severely than smaller ones.

QWK is particularly valuable in the evaluation of automated scoring for assessing how well LLM-generated scores align with those from human raters. A QWK score close to 1 indicates strong agreement, suggesting comparable scores, while a lower QWK score signals greater divergence. This metric is widely used in the evaluation of automated scoring.

\subsubsection{Normalized Mutual Information}
NMI is an information-theoretic metric that can be used to measure the mutual dependence between two discrete distributions. It is derived from mutual information (MI), which quantifies the information shared between two random variables \citep{scikit-learn}. Given two discrete score distributions, \(X\) and \(Y\), the MI is defined as follows:
\begin{equation}
\mathrm{MI}(X, Y) = \sum_{x \in X} \sum_{y \in Y} P(x, y) \log \frac{P(x, y)}{P(x)P(y)}
\end{equation}
where \(P(x,y)\) is the joint probability distribution, and \(P(x)\) and \(P(y)\) are the marginal probabilities. To make MI comparable across different distributions, it is normalized as
\begin{equation}
\mathrm{NMI}(X, Y) = \frac{2 \times \mathrm{MI}(X, Y)}{H(X) + H(Y)}
\end{equation}
where \(H(X)\) and \(H(Y)\) are the entropies of the respective distributions,
\begin{equation}
H(X) = - \sum_{x \in X} P(x) \log P(x), \quad H(Y) = - \sum_{y \in Y} P(y) \log P(y)
\end{equation}
NMI values range from 0 to 1, where 1 indicates a perfect match between distributions and 0 indicates that there is no mutual information. This makes it a useful metric for evaluating the agreement between different scoring methods.

In automated scoring, NMI can be used to indicate how well LLM-generated scores align with human ratings. An NMI value close to 1 suggests strong alignment with human scoring patterns, ensuring scoring consistency. A lower NMI value indicates discrepancies between the two scoring systems, suggesting that there may be inconsistencies. In this study, scores assigned by human raters are compared with those assigned by each LLM.

\subsubsection{Cosine Similarity}

Cosine similarity is widely used in NLP and information retrieval for quantifying the similarity between two vectors. It measures the cosine of the angle between two nonzero vectors in a multidimensional space, indicating how similar they are in direction regardless of magnitude \citep{Manning2008}. Given two vectors \(X\) and \(Y\), their cosine similarity is computed as
\begin{equation}
\cos(\theta) = \frac{X \cdot Y}{||X|| ||Y||} = \frac{\sum_{i=1}^{n} X_i Y_i}{\sqrt{\sum_{i=1}^{n} X_i^2} \sqrt{\sum_{i=1}^{n} Y_i^2}}
\end{equation}
where \(X \cdot Y\) is the dot product of the two vectors, and \(\|X\|\) and \(\|Y\|\) denote their Euclidean norms. The resulting similarity scores range from $-1$ to $1$, where $1$ indicates identical vectors, $0$ suggests no similarity, and $-1$ represents complete dissimilarity in opposite directions.

In this study, cosine similarity is used to compare the rationales of LLMs and human raters. As described in the Text Embedding section, LLM-generated rationales and human raters' rationales are represented as high-dimensional embedding vectors. A high cosine similarity value between these vectors implies strong agreement between the human and LLM rationales, while a lower value indicates divergence in rating rationales.

\subsubsection{Principal Component Analysis}

PCA is a widely used statistical technique for reducing the dimensionality of high-dimensional datasets while preserving as much variance as possible. It achieves this by transforming the original correlated variables into a new set of uncorrelated variables called principal components, which are ranked by the amount of variance they explain in the data. Given the embeddings for the rationales provided by LLM and human raters, PCA can be used to find a set of orthonormal basis vectors that maximize variance in the data projection. Mathematically, PCA is performed by computing the eigenvectors and eigenvalues of the covariance matrix of the input data \citep{mackiewicz1993principal}.

Let \( X \) be a dataset with mean-centered features. The covariance matrix of \( X \) is given by
\begin{equation}
C = \frac{1}{n} X^T X
\end{equation}
PCA is then used to solve the eigenvalue problem
\begin{equation}
C v = \lambda v
\end{equation}
where \( v \) represents the eigenvectors (principal components) and \(\lambda\) are the corresponding eigenvalues, which indicate the amount of variance captured by each component. The principal components are ranked by descending eigenvalue, and the top-\( k \) components are selected to form a reduced representation of the data.

In this study, PCA is used to help analyze the high-dimensional embeddings extracted from textual data generated by human raters and LLMs. To do this, the rationale texts are converted into high-dimensional vectors using text embedding. PCA is then applied to these embeddings, and the results are visualized in two-dimensional space for easier interpretation. Data points that are closer together in the transformed space indicate similar embedding representations, suggesting a high degree of similarity in scoring rationales; conversely, points that are farther apart indicate greater differences in the text.

\section{Results}
The results of this study are organized into three parts: considering the score consistency between LLMs and two human raters (R1 and R2), evaluation of rationale similarity using SBERT embeddings and cosine similarity, and visualization of rationale embeddings through PCA. Since the rationale of R2 was not available, only the rationale of R1 is compared with the rationales generated by the various LLMs.

Table~\ref{tab:stat_summary} presents summary statistics relating to the consistency between the scores assigned by each LLM and human raters~R1 and R2. Two metrics were used to evaluate this: QWK and NMI. Among all studied LLMs, GPT-4o demonstrated the highest level of agreement with both R1 and R2, achieving QWK scores of 0.8646 and 0.8976, respectively. It also attained the highest NMI value with R2 (0.6581) and a high NMI with R1 (0.5430). Claude~3.5 Sonnet also performed well, with QWK scores of 0.8041 (R1) and 0.8964 (R2) and NMI scores of 0.5057 (R1) and 0.6515 (R2). GPT-4 achieved QWK values of 0.8270 (R1) and 0.8339 (R2), alongside NMI scores of 0.5621 (R1) and 0.5017 (R2). In contrast, Gemini~1.5 had the lowest agreement, with QWK values of 0.4355 (R1) and 0.4631 (R2) and NMI scores of 0.2071 (R1) and 0.2974 (R2).

\begin{table}[htbp]
\centering
\caption{Summary Statistics of Scoring Consistency between LLMs and Human Raters}
\label{tab:stat_summary}
\begin{tabular}{lrrrrrrrr}
\toprule Model & Mean & Std. Dev. & NMI (R1) & NMI (R2) & QWK (R1) & QWK (R2) \\
\midrule GPT-3.5           & 3.13 & 1.0080 & 0.4922 & 0.4271 & 0.6322 & 0.5601 \\
GPT-4.0           & 3.93 & 1.2299 & 0.5621 & 0.5017 & 0.8270 & 0.8339 \\
GPT-4o            & 3.77 & 1.2507 & 0.5430 & 0.6581 & 0.8646 & 0.8976 \\
Gemini~1.5        & 3.60 & 0.6747 & 0.2071 & 0.2974 & 0.4355 & 0.4631 \\
Claude~3.5 Sonnet & 3.87 & 1.3830 & 0.5057 & 0.6515 & 0.8041 & 0.8964 \\
OpenAI~o1         & 3.10 & 1.0289 & 0.4141 & 0.4793 & 0.6565 & 0.5842 \\
Gemini 2.0        & 3.33 & 0.8841 & 0.4848 & 0.6072 & 0.7158 & 0.6186 \\
R1 (Human)        & 3.73 & 1.1121 & --     & 0.6772 & --     & 0.8735 \\
R2 (Human)        & 4.07 & 1.1725 & 0.6772 & --     & 0.8735 & -- \\
\bottomrule
\end{tabular}
\end{table}

To evaluate the textual similarity between the rationales generated by the LLMs and a human rater (R1), cosine similarity was computed using SBERT embeddings. Tables~\ref{tab:cosine_similarity_sentence_bert_same}--\ref{tab:cosine_similarity_sentence_bert_diff_2} summarize the results, conditional on absolute score differences of~0, 1, and 2.

When the model score matched the R1 score (i.e., an absolute score difference of~0; Table~\ref{tab:cosine_similarity_sentence_bert_same}), the highest mean cosine similarity was achieved by OpenAI~o1 (0.5972), followed by Claude~3.5 Sonnet (0.5744) and GPT-3.5 (0.4595). In terms of maximum similarity, OpenAI~o1 again led with 0.6857, while Claude~3.5 Sonnet (0.6797) and GPT-3.5 (0.6706) also performed strongly. The lowest minimum similarity was observed for GPT-4.0 (0.2086), while Claude~3.5 Sonnet showed the narrowest range (with a minimum of 0.4366), indicating more consistent alignment with human rationales.

When the absolute score difference increased to~1 (Table~\ref{tab:cosine_similarity_sentence_bert_diff_1}), cosine similarity values generally increased. Claude~3.5 Sonnet maintained the highest mean similarity of 0.6224, and this was followed by OpenAI~o1 at 0.6101. GPT-3.5, GPT-4.0, and GPT-4o yielded mean values of 0.5444, 0.4787, and 0.4570, respectively. Notably, OpenAI~o1 achieved the highest maximum similarity across all models and conditions (0.8711), while GPT-4.0 had the lowest minimum similarity again (0.2383), indicating a wider range of alignment quality.

For the two-point score-difference condition (Table~\ref{tab:cosine_similarity_sentence_bert_diff_2}), only a subset of models could be considered because GPT-4.0 and GPT-4o did not yield any two-point score differences with R1, and they were thus not included. Claude~3.5 Sonnet again exhibited the highest mean similarity (0.5721), with a narrow range between maximum (0.5829) and minimum (0.5613), suggesting a stable similarity with human-rater rationales. OpenAI~o1 followed with a mean of 0.5151 and a maximum of 0.6225. In contrast, GPT-3.5 and Gemini~1.5 produced lower mean values of 0.3731 and 0.3924, respectively, with GPT-3.5 also showing the lowest minimum similarity (0.2325).

\begin{table}[htbp]
\centering
\caption{Cosine Similarity between LLM- and Human-Generated Rationales for Matched Scores}
\label{tab:cosine_similarity_sentence_bert_same}
\begin{tabular}{lrrrrr}
\toprule Model & Max & Min & Mean & Std. Dev. & Count \\
\midrule A1 – GPT-3.5 & 0.6706 & 0.2612 & 0.4595 & 0.1128 & 14 \\
A2 – GPT-4.0 & 0.6333 & 0.2086 & 0.4333 & 0.1087 & 16 \\
A3 – GPT-4o & 0.6689 & 0.2779 & 0.4521 & 0.1109 & 19 \\
A4 – Gemini~1.5 & 0.5977 & 0.2630 & 0.4457 & 0.0941 & 12 \\
A5 – Claude~3.5 Sonnet & 0.6797 & 0.4366 & 0.5744 & 0.0690 & 18 \\
A6 – OpenAI~o1 & 0.6857 & 0.3977 & 0.5972 & 0.0780 & 15 \\
A7 – Gemini 2.0 & 0.6464 & 0.1996 & 0.4506 & 0.1307 & 14 \\
\bottomrule
\end{tabular}
\end{table}

\begin{table}[htbp]
\centering
\caption{Cosine Similarity between LLM- and R1-Generated Rationales for One-Score Differences}
\label{tab:cosine_similarity_sentence_bert_diff_1}
\begin{tabular}{lrrrrr}
\toprule Model & Max & Min & Mean & Std. Dev. & Count \\
\midrule A1 – GPT-3.5 & 0.6798 & 0.3742 & 0.5444 & 0.0839 & 12 \\
A2 – GPT-4.0 & 0.6801 & 0.2383 & 0.4787 & 0.1249 & 14 \\
A3 – GPT-4o & 0.5958 & 0.3531 & 0.4570 & 0.0796 & 11 \\
A4 – Gemini~1.5 & 0.5625 & 0.3069 & 0.4212 & 0.0741 & 12 \\
A5 – Claude~3.5 Sonnet & 0.7147 & 0.4032 & 0.6224 & 0.0852 & 10 \\
A6 – OpenAI~o1 & 0.8711 & 0.4270 & 0.6101 & 0.1460 & 11 \\
A7 – Gemini 2.0 & 0.5813 & 0.5813 & 0.5813 & 0.0000 & 1 \\
\bottomrule
\end{tabular}
\end{table}

\begin{table}[htbp]
\centering
\caption{Cosine Similarity between LLM- and R1-Generated Rationales for Two-Score Differences}
\label{tab:cosine_similarity_sentence_bert_diff_2}
\begin{tabular}{lrrrrr}
\toprule Model & Max & Min & Mean & Std. Dev. & Count \\
\midrule A1 – GPT-3.5 & 0.5301 & 0.2325 & 0.3731 & 0.1057 & 4 \\
A4 – Gemini~1.5 & 0.4829 & 0.3034 & 0.3924 & 0.0839 & 4 \\
A5 – Claude~3.5 Sonnet & 0.5829 & 0.5613 & 0.5721 & 0.0107 & 2 \\
A6 – OpenAI~o1 & 0.6225 & 0.4252 & 0.5151 & 0.0712 & 4 \\
\bottomrule
\end{tabular}
\end{table}

\begin{figure}[htbp]
    \centering
    \includegraphics[width=\textwidth]{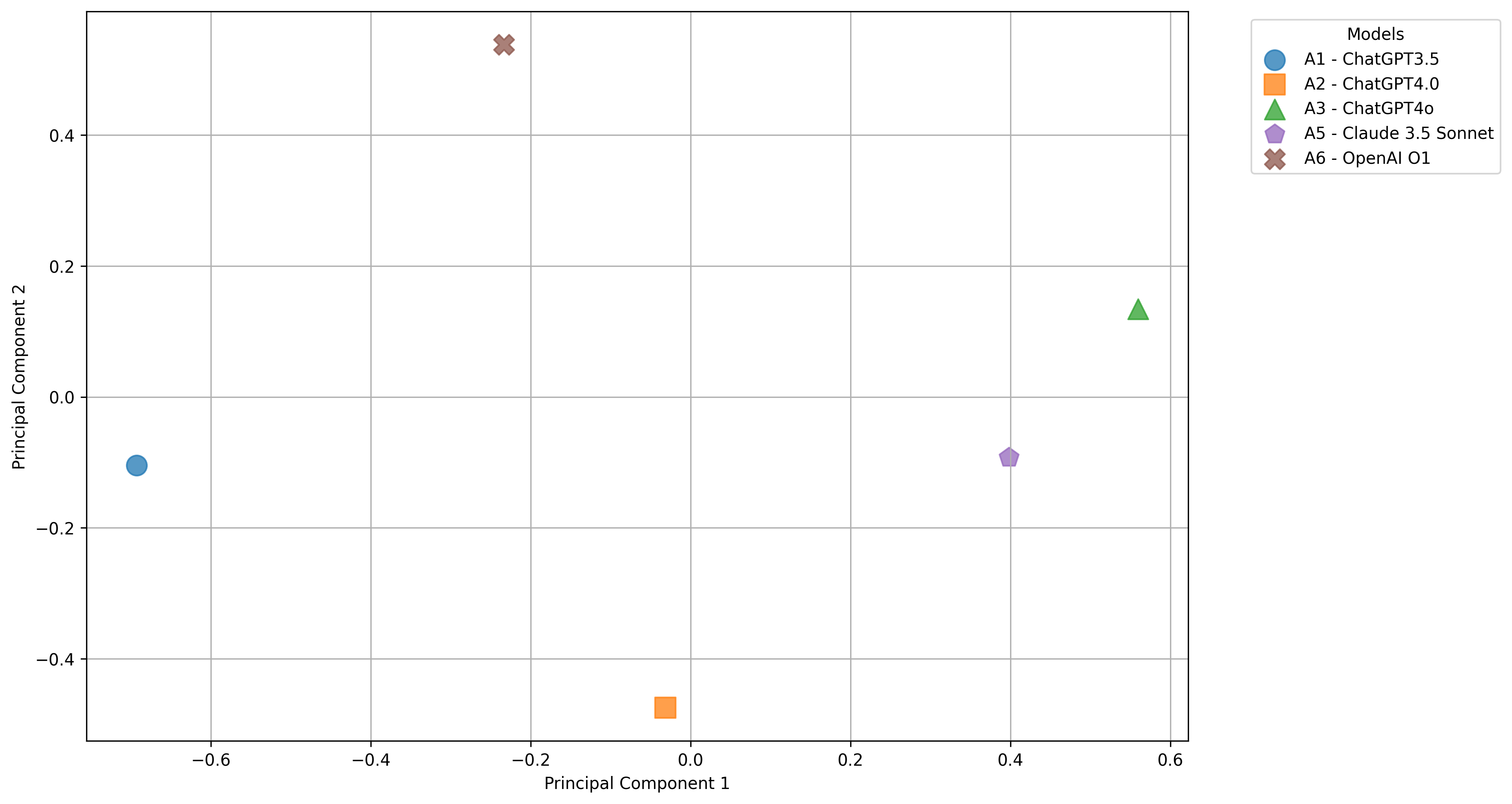}
    \caption{Visualization of PCA Score 1 Distribution}
    \label{fig:pca1}
\end{figure}

\begin{figure}[htbp]
    \centering
    \includegraphics[width=\textwidth]{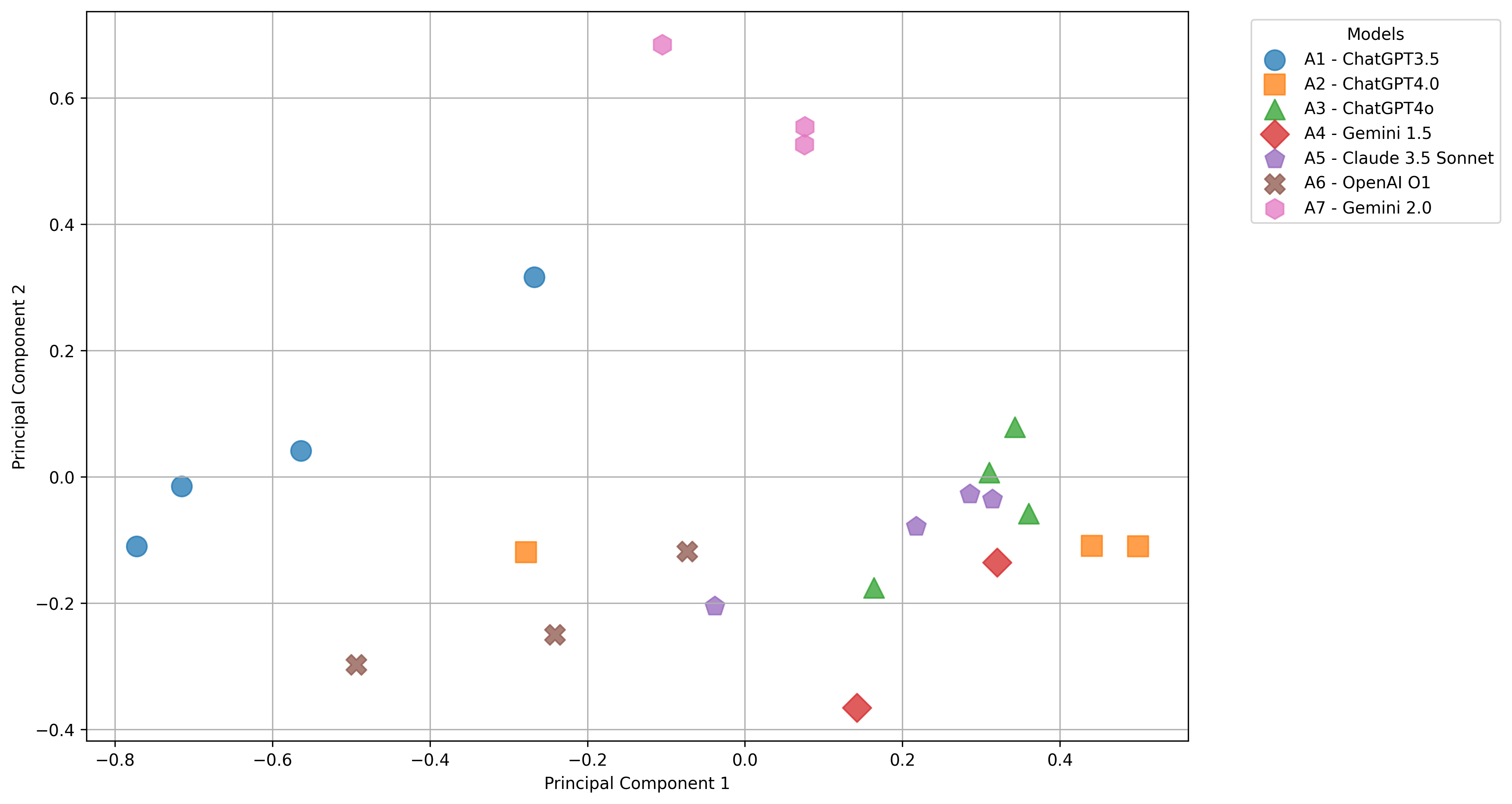}
    \caption{Visualization of PCA Score 2 Distribution}
    \label{fig:pca2}
\end{figure}

\begin{figure}[htbp]
    \centering
    \includegraphics[width=\textwidth]{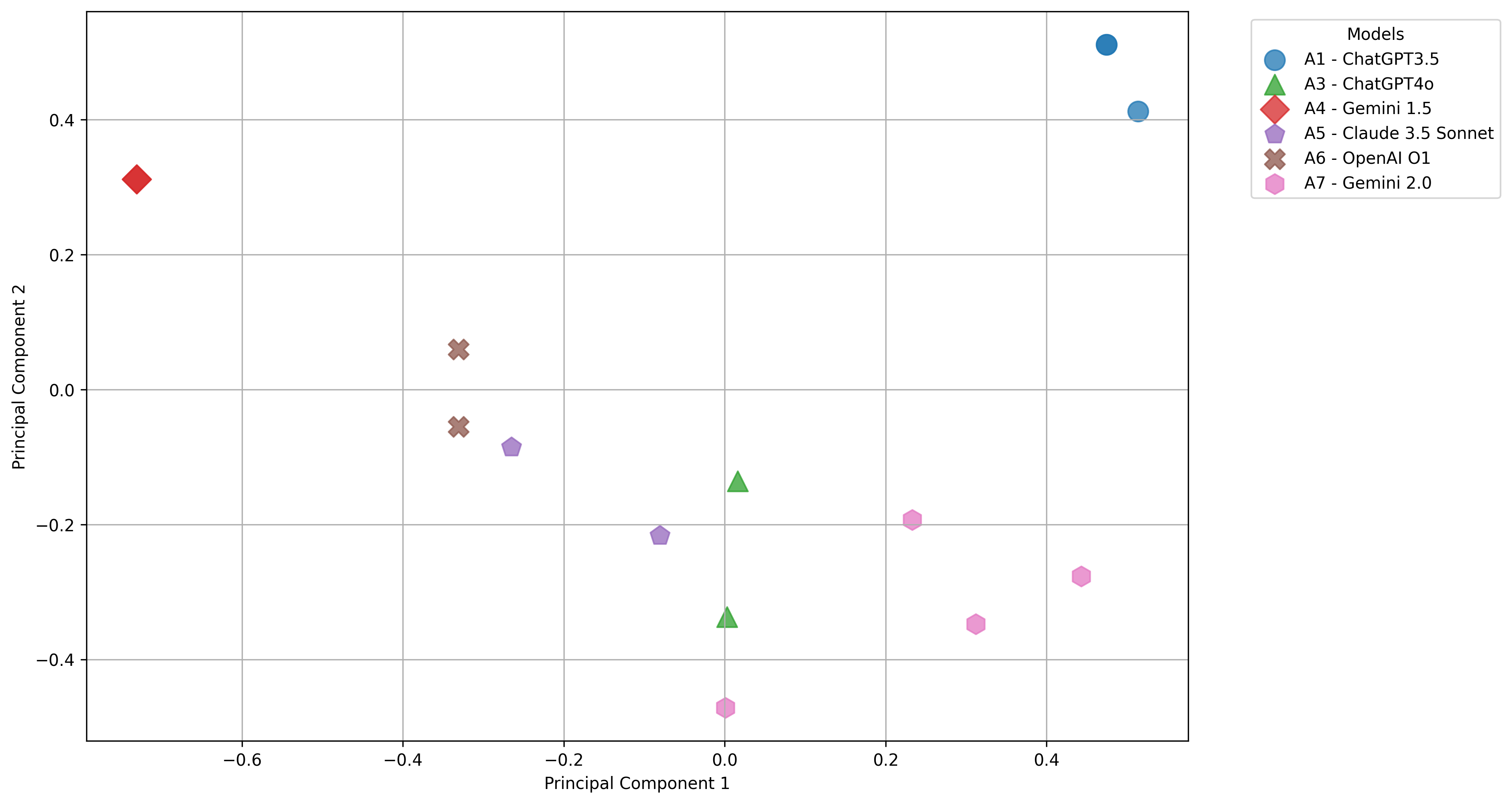}
    \caption{Visualization of PCA Score 3 Distribution}
    \label{fig:pca3}
\end{figure}

\begin{figure}[htbp]
    \centering
    \includegraphics[width=\textwidth]{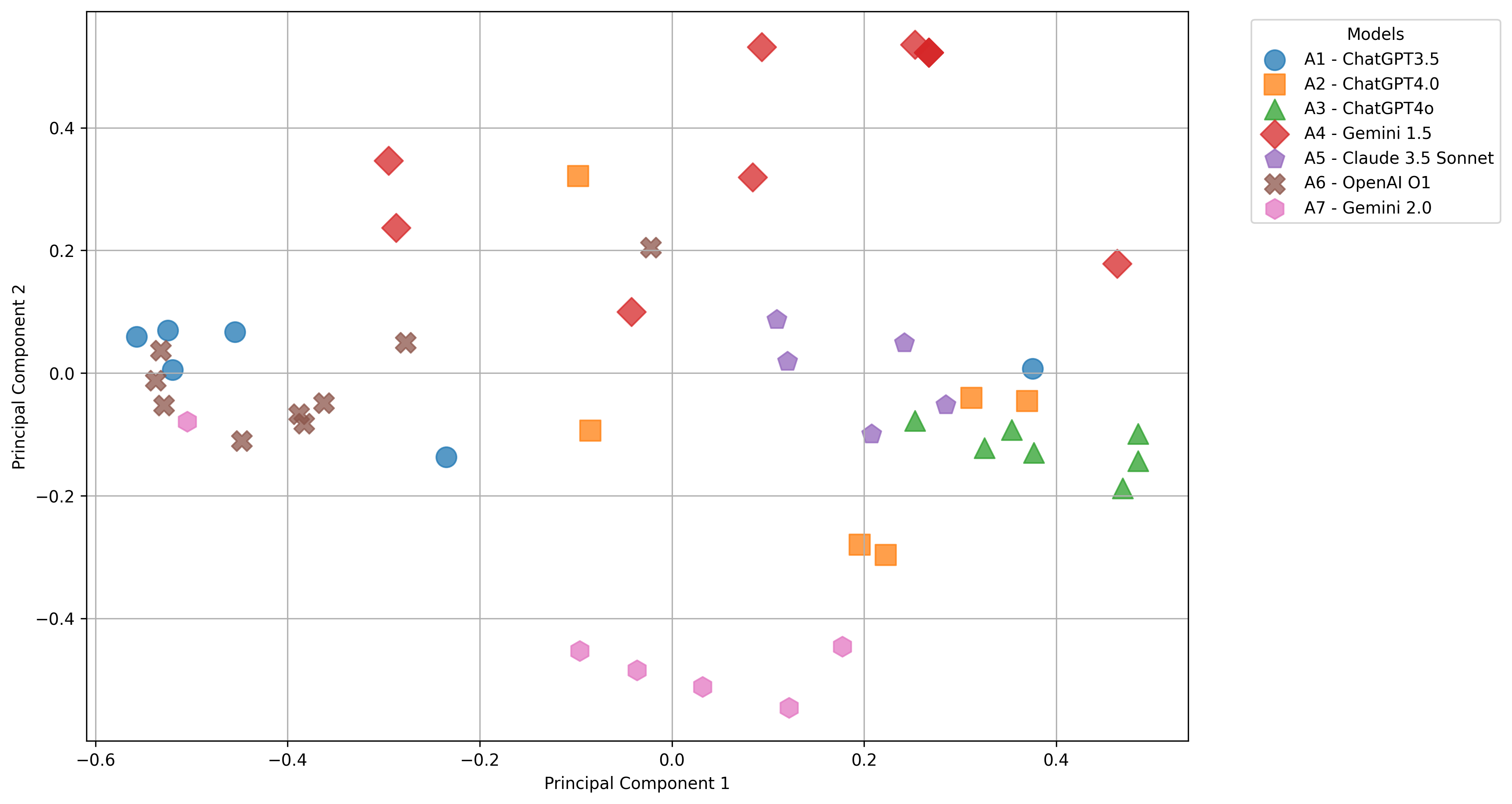}
    \caption{Visualization of PCA Score 4 Distribution}
    \label{fig:pca4}
\end{figure}

\begin{figure}[htbp]
    \centering
    \includegraphics[width=\textwidth]{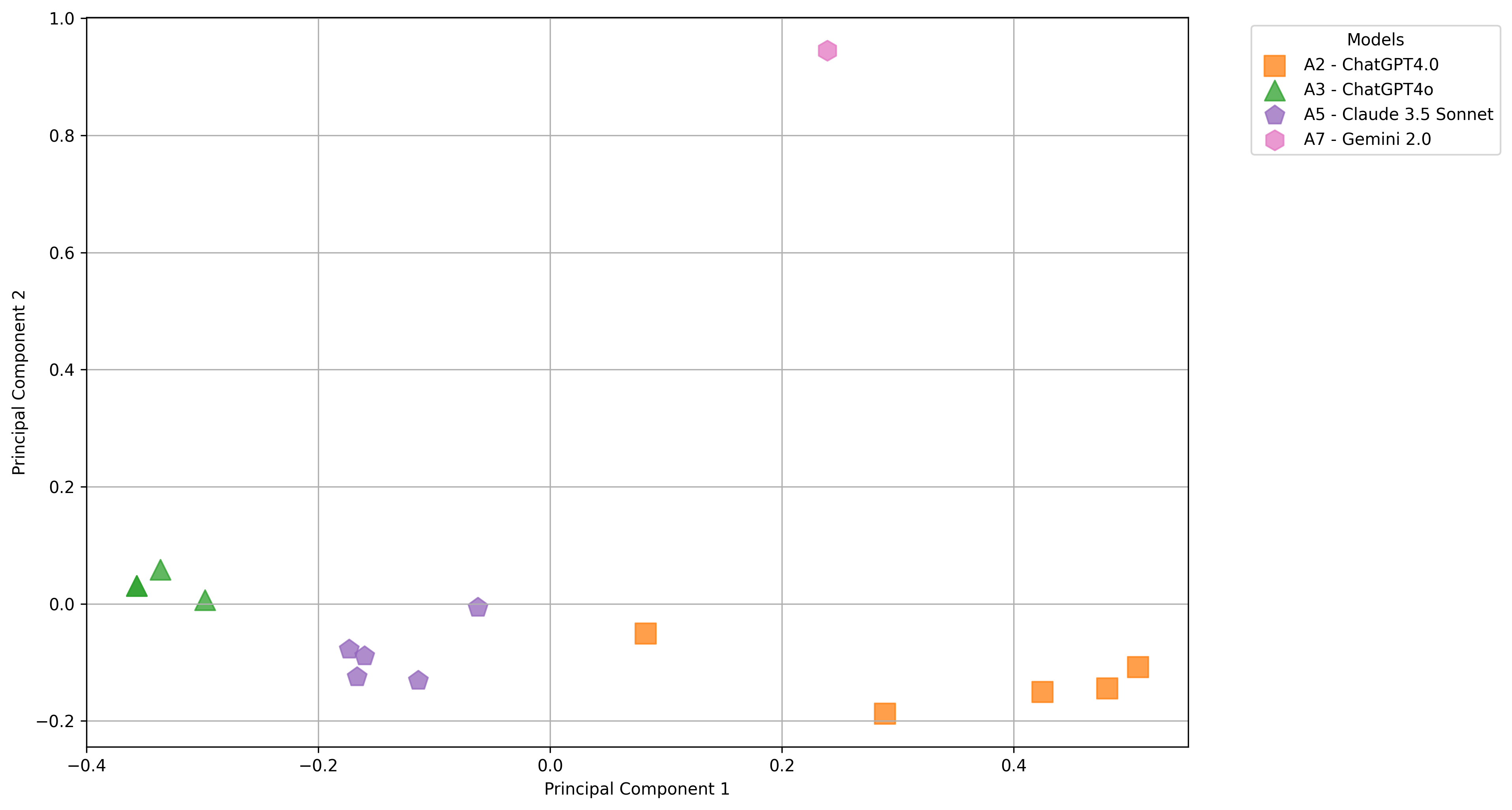}
    \caption{Visualization of PCA Score 5 Distribution}
    \label{fig:pca5}
\end{figure}

\begin{figure}[htbp]
    \centering
    \includegraphics[width=\textwidth]{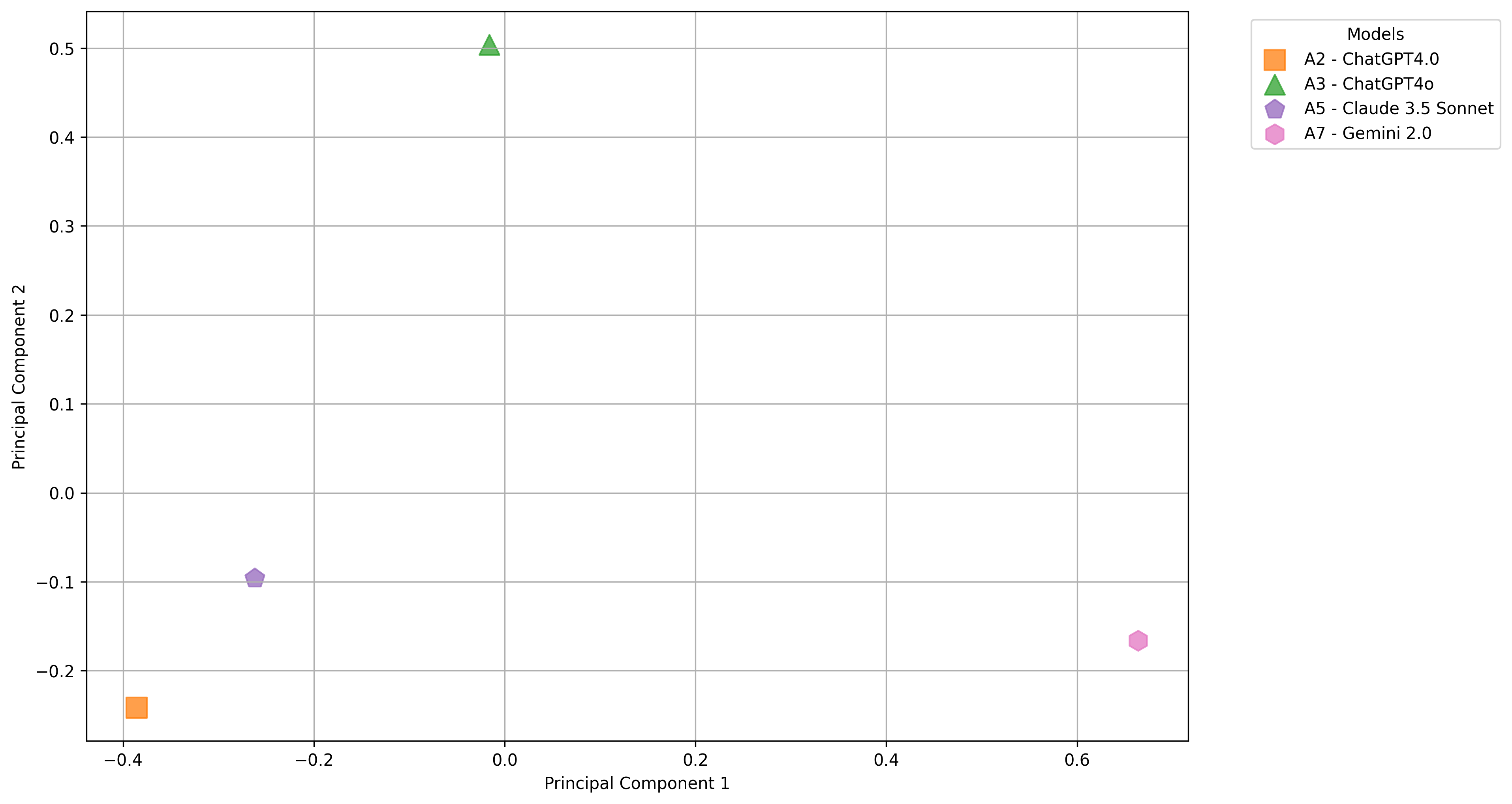}
    \caption{Visualization of PCA Score 6 Distribution}
    \label{fig:pca6}
\end{figure}

To better understand how rationale embeddings vary across LLMs, PCA was applied to the SBERT embeddings for dimensionality reduction to two dimensions. Figures~\ref{fig:pca1}--\ref{fig:pca6} present PCA plots for specific scores (1 through 6, respectively), including only rationale embeddings where LLM scores matched those of R1. In low-score categories (e.g., score~1, Figure~\ref{fig:pca1}), the rationales produced by the LLMs appear more dispersed, indicating greater semantic variability. As the scores increase, the LLM-generated rationales show tighter clustering, reflecting higher semantic similarity. For example, in score~5 (Figure~\ref{fig:pca5}) and score~6 (Figure~\ref{fig:pca6}), embeddings from GPT-4o, Claude~3.5 Sonnet, and OpenAI~o1 form compact clusters, suggesting stronger alignment in the content of the rationales. These visual trends support the cosine-similarity findings: LLMs that align more closely with human scores tend to produce more semantically consistent rationales.

\begin{figure}[htbp]
    \centering
    \includegraphics[width=0.75\textwidth]{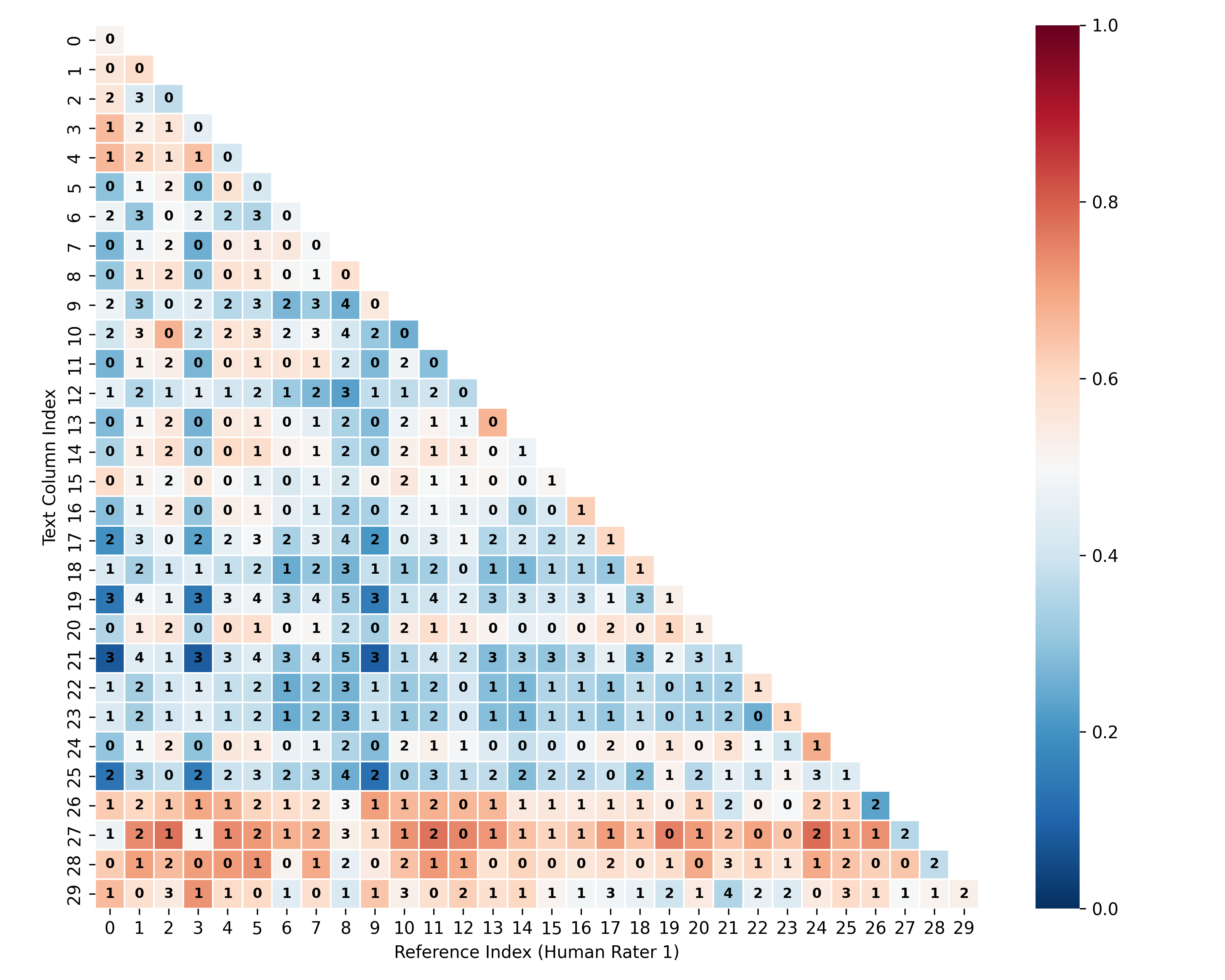}
    \caption{Heatmap of Cosine Similarity for GPT-3.5}
    \label{fig:cs3.5}
\end{figure}

\begin{figure}[htbp]
    \centering
    \includegraphics[width=0.75\textwidth]{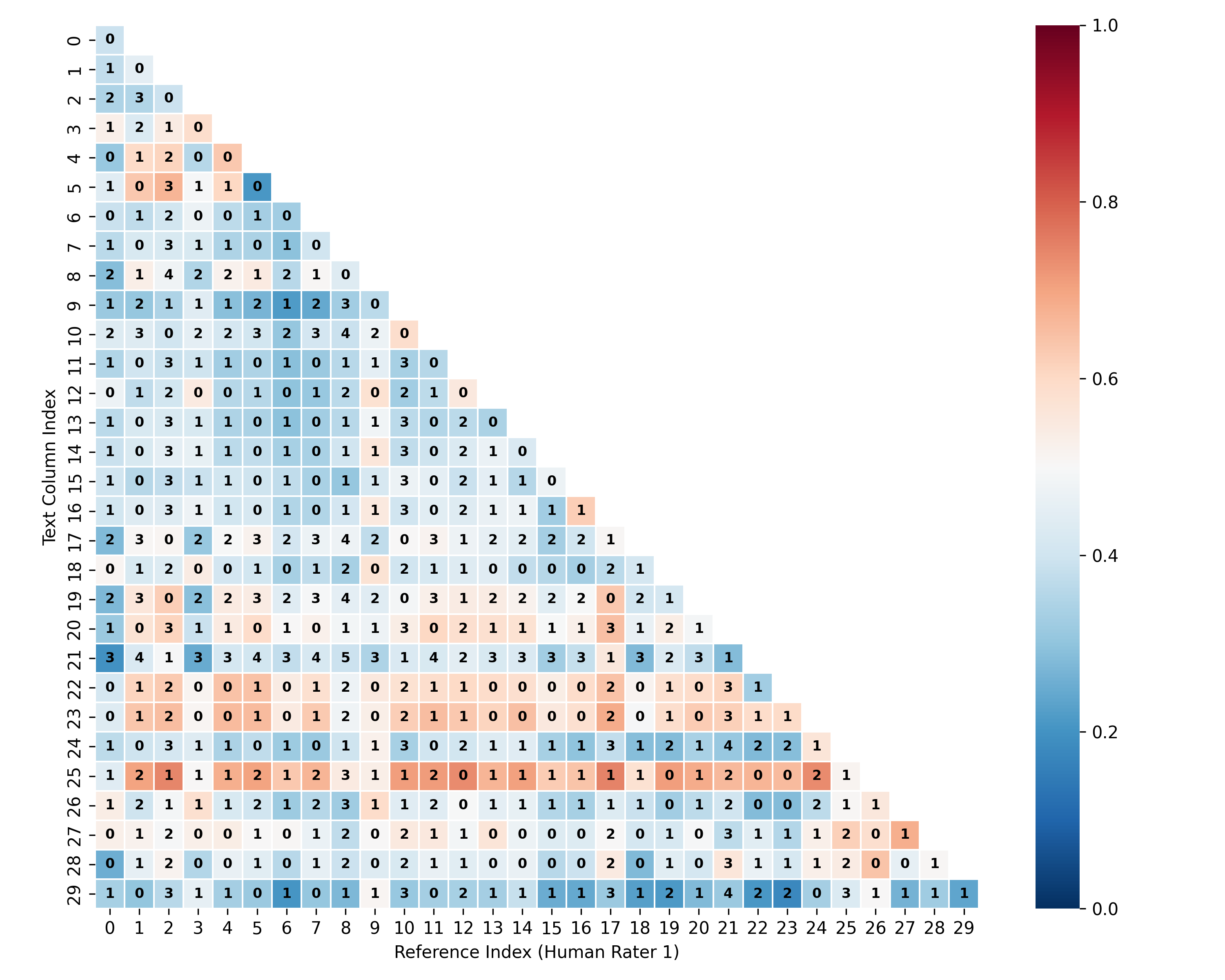}
    \caption{Heatmap of Cosine Similarity for GPT-4.0}
    \label{fig:cs4.0}
\end{figure}

\begin{figure}[htbp]
    \centering
    \includegraphics[width=0.75\textwidth]{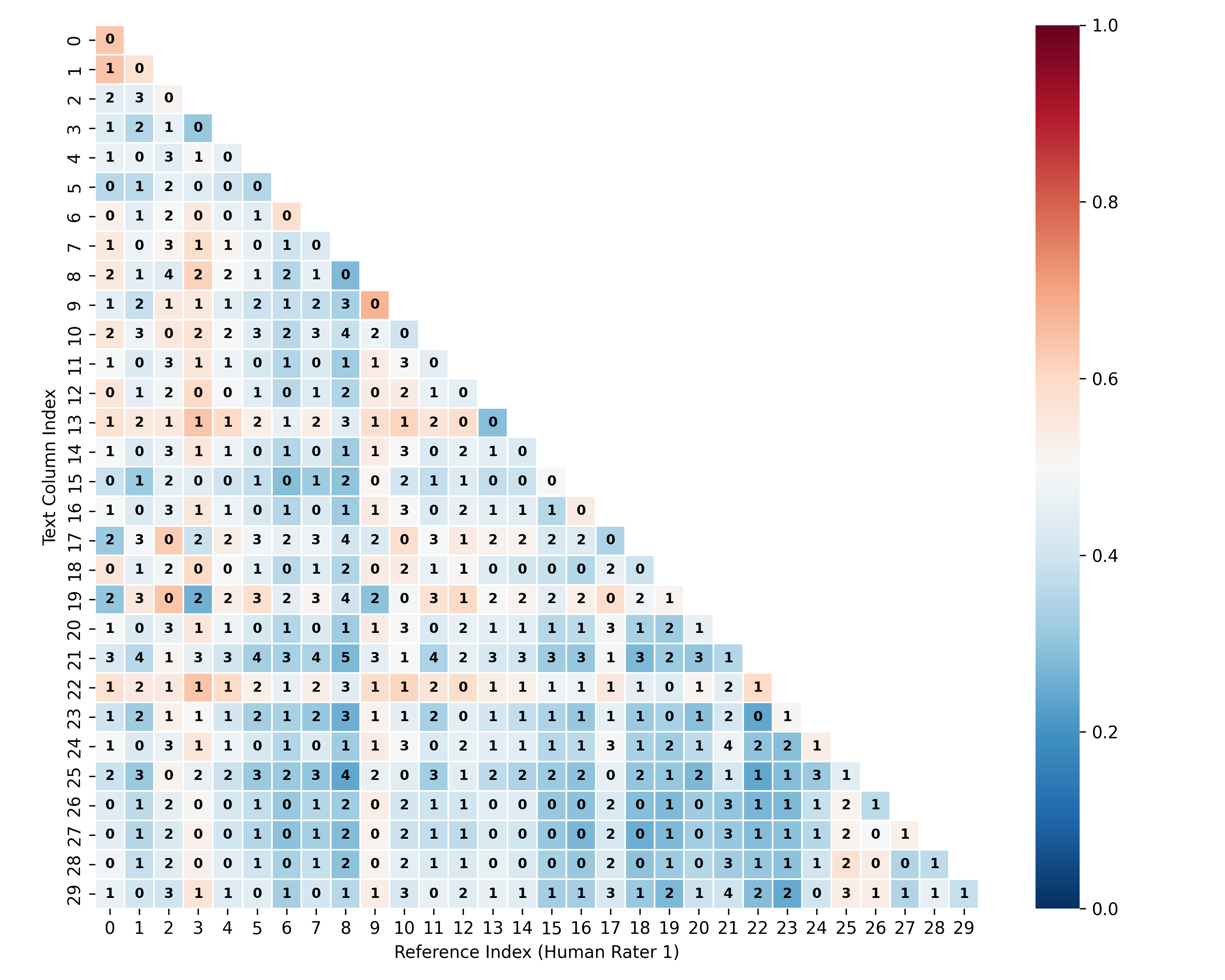}
    \caption{Heatmap of Cosine Similarity for GPT-4o}
    \label{fig:cs4O}
\end{figure}

\begin{figure}[htbp]
    \centering
    \includegraphics[width=0.75\textwidth]{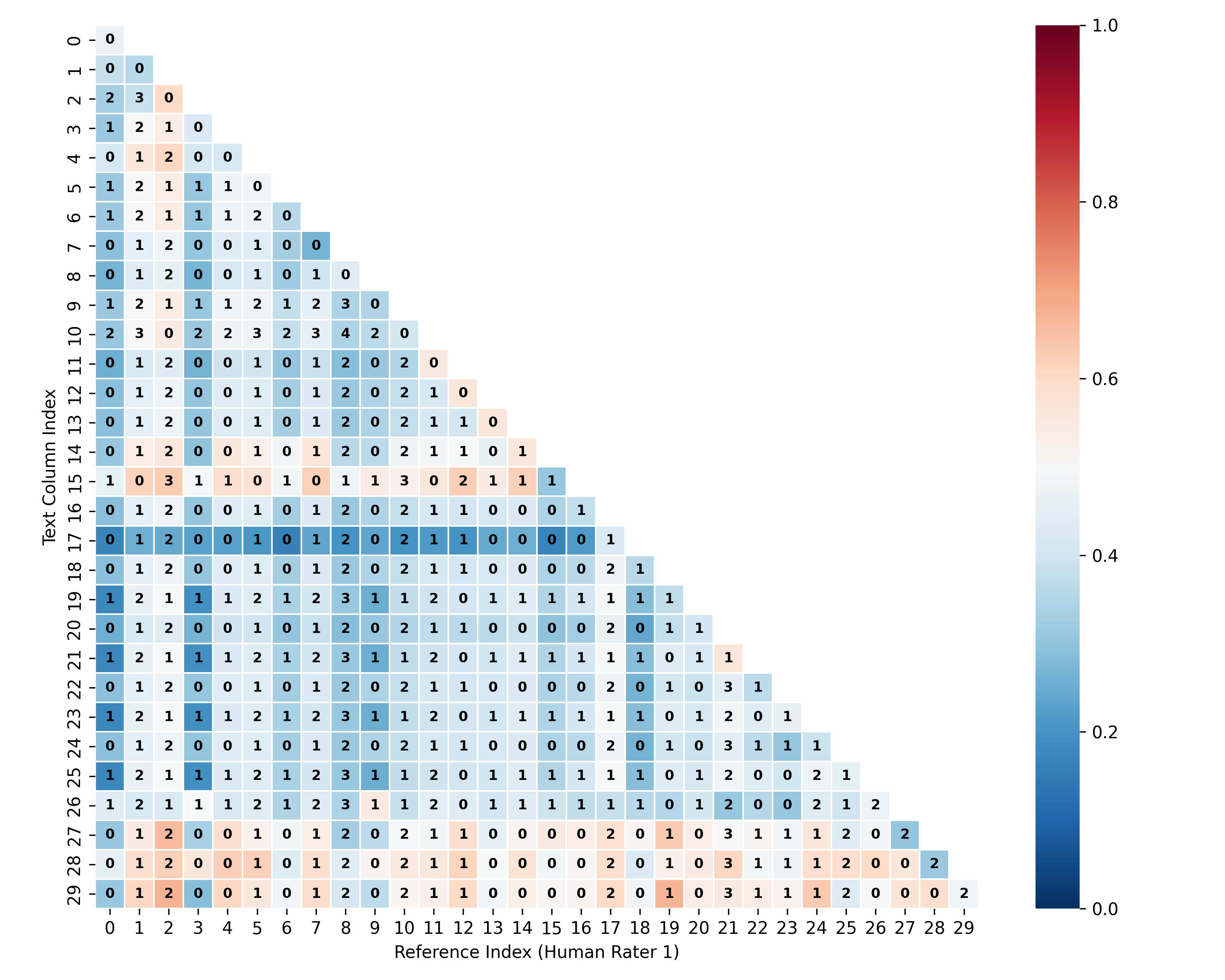}
    \caption{Heatmap of Cosine Similarity for Gemini}
    \label{fig:csGemini}
\end{figure}

\begin{figure}[htbp]
    \centering
    \includegraphics[width=0.75\textwidth]{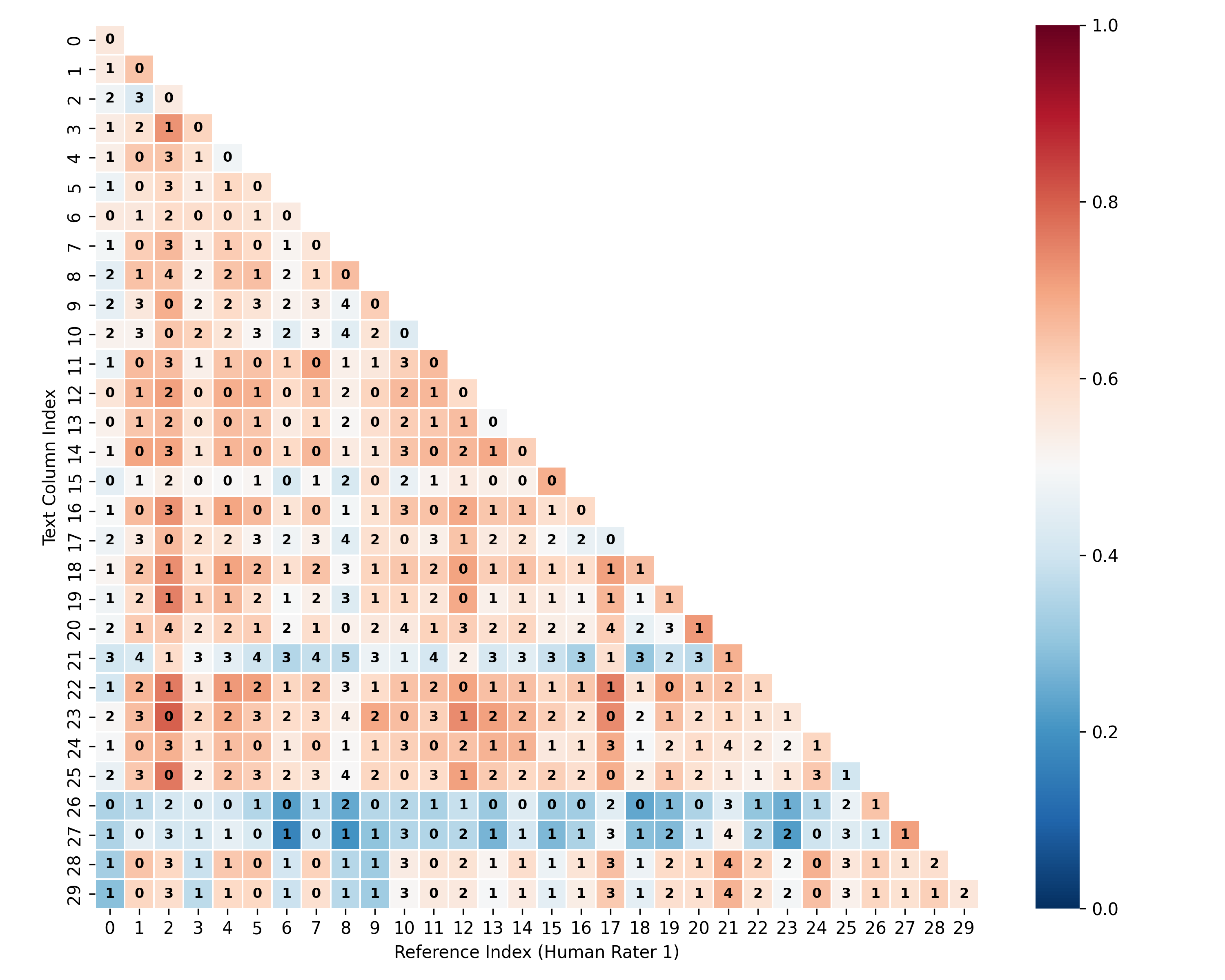}
    \caption{Heatmap of Cosine Similarity for Claude~3.5 Sonnet}
    \label{fig:csClaude}
\end{figure}

\begin{figure}[htbp]
    \centering
    \includegraphics[width=0.75\textwidth]{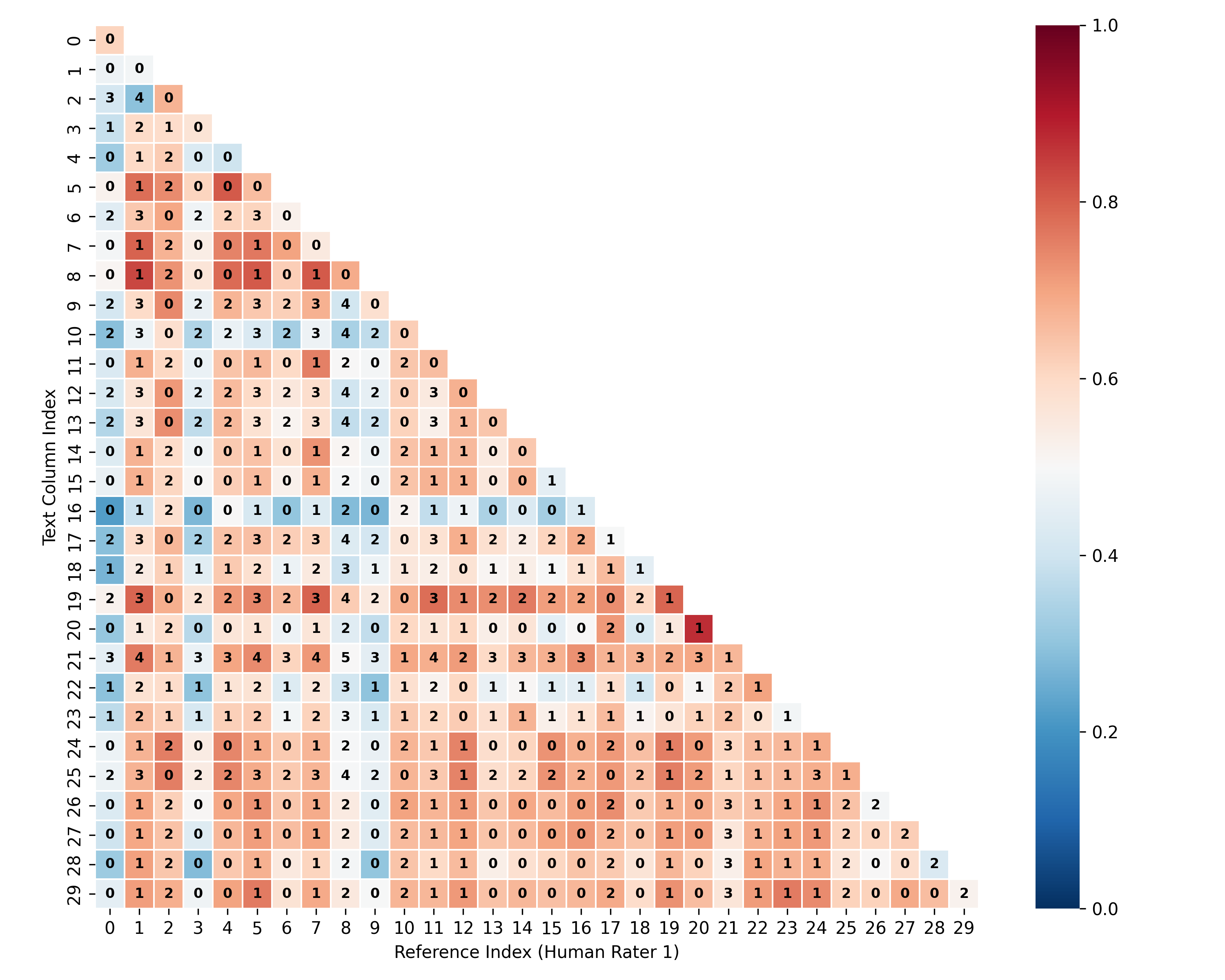}
    \caption{Heatmap of Cosine Similarity for OpenAI O1}
    \label{fig:cso1}
\end{figure}

\begin{figure}[htbp]
    \centering
    \includegraphics[width=0.75\textwidth]{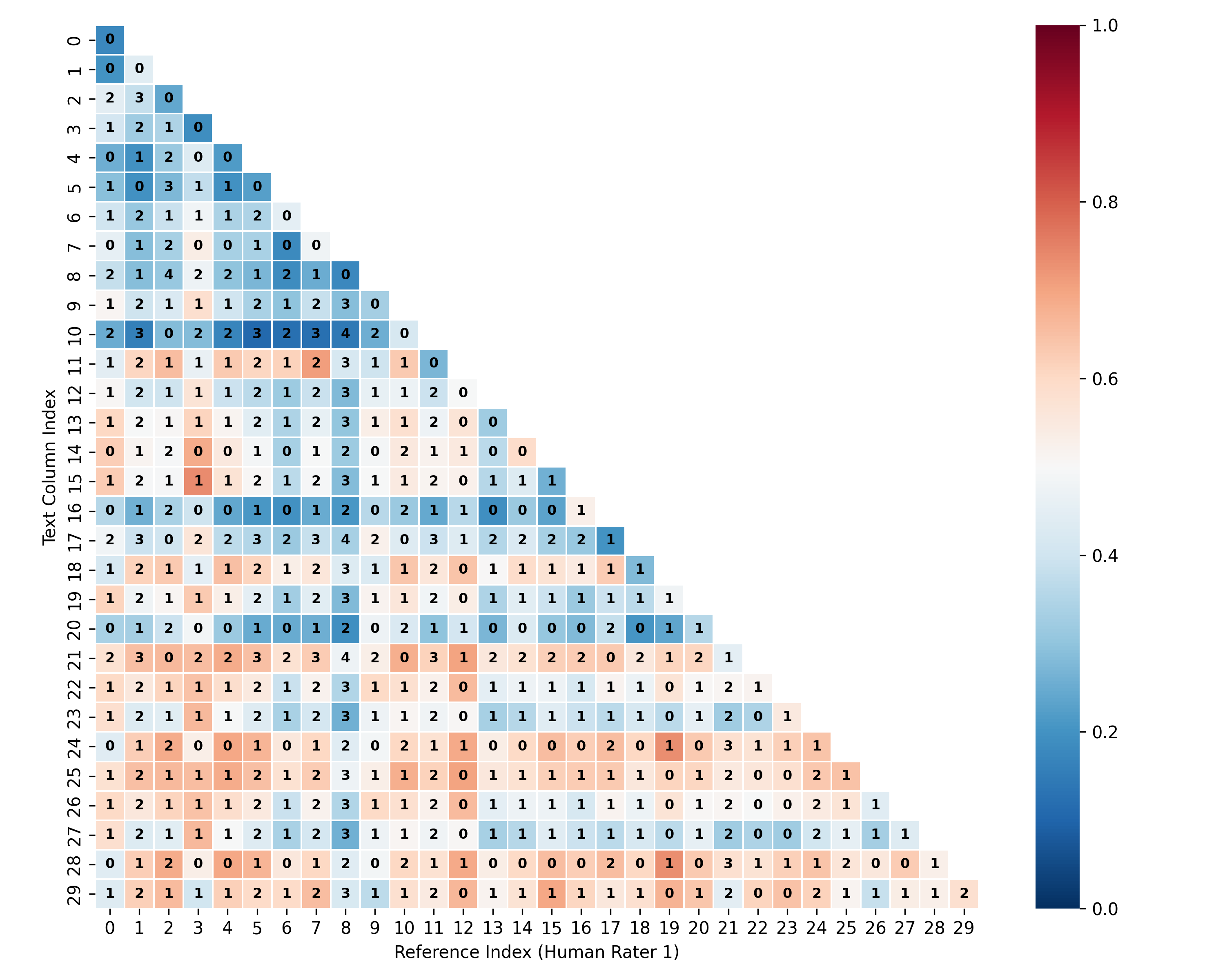}
    \caption{Heatmap of Cosine Similarity for Gemini 2.0}
    \label{fig:csgemini2.0}
\end{figure}

Figures~\ref{fig:cs3.5}--\ref{fig:csgemini2.0} show heatmaps of cosine similarity between rationale embeddings generated by various LLMs and those provided by the reference human rater (R1). Each heatmap cell represents the semantic similarity of a rationale pair based on SBERT embeddings, and the overlaid numeric labels indicate absolute score differences. Moderate semantic alignment is evident in Figure~\ref{fig:cs3.5} (GPT-3.5), but there are scattered darker blue cells and frequent score mismatches, reflecting partial consistency. Figure~\ref{fig:cs4.0} (GPT-4.0) shows improved alignment, featuring larger contiguous regions of high similarity, but some inconsistencies remain under one-point score differences. Figure~\ref{fig:cs4O} (GPT-4o) demonstrates strong and consistent rationale similarity with minimal score gaps, visually confirming its leading performance for both QWK and NMI. In contrast, Figure~\ref{fig:csGemini} (Gemini~1.5) reveals weaker alignment, with lighter cell colors and frequent score differences, indicating lower semantic similarity and reliability. Figure~\ref{fig:csClaude} (Claude~3.5 Sonnet) shows strong rationale alignment and minimal score differences, suggesting that it can closely simulate human evaluative reasoning. Figure~\ref{fig:cso1} (OpenAI~o1) demonstrates high similarity when scores match, though there is more variability when score discrepancies arise. Lastly, Figure~\ref{fig:csgemini2.0} (Gemini~2.0) shows some improvement over Gemini~1.5, with more medium- and high-similarity regions, though moderate divergence remains when scores differ. Overall, models with stronger score agreement, such as GPT-4o and Claude~3.5 Sonnet, tend to produce rationales that are more semantically aligned with those written by human raters.

\section{Summary and Discussion}

This study compares the scores and rationales generated by LLMs with those of human raters in the context of AP Chinese writing assessments. The results suggest that newer versions of LLMs from each model family, such as GPT-4o and Claude~3.5 Sonnet, demonstrate superior score consistency with human raters in terms of QWK and NMI, as well producing rationales with stronger cosine similarity.

One notable finding is that the scoring consistency aligns with the semantic similarity in the rationales. When scores produced by LLMs matched those of the human raters, the cosine-similarity values were consistently higher. This supports the notion that high-performing LLMs not only produce scores that align well with human evaluations but also generate rationales that closely mirror those of human raters. This has significant practical implications for the trustworthiness and pedagogical integration of LLMs into formative writing assessments, where valid reasoning is as important as accurate scoring.

Nonetheless, notable differences were observed among LLMs. While GPT-4o demonstrated both high score consistency and high rationale similarity, models such as Gemini~1.5 had weaker performance across all evaluation criteria. These disparities may stem from differences in architecture, training objectives, or alignment strategies. As LLM-assisted writing evaluation becomes more common, the observed variability across models calls for caution: different LLMs can produce noticeably different results in automated essay scoring. Empirical validation is therefore essential before specific LLMs are deployed for high-stakes scoring tasks.

The PCA plots further support these findings, revealing tighter clustering of rationale embeddings for LLMs whose scores aligned with those of~R1. These patterns were particularly pronounced at higher score levels (5 and 6), suggesting that alignment with human raters encompasses not just scoring accuracy but also similarity in underlying reasoning. Embedding-based dimensionality-reduction techniques such as PCA offer an important lens into the otherwise opaque reasoning processes of LLMs, helping to reveal not just what predictions LLMs make but how those predictions are justified.

These findings are by no means intended to be generalized to other items or content domains. Rather, this study demonstrates methods that can be used for evaluating LLMs in terms of both score consistency and rationale similarity. The findings underscore the need for further research into the interpretability and reliability of LLM-generated rationales, especially in high-stakes assessment contexts. Moreover, examining how LLMs behave under inconsistencies in scoring could further illuminate their limitations and inform the design of feedback systems that support student learning. Future work may consider the use of chain-of-thought reasoning when prompting LLMs for automated scoring.

\section{Conclusion}

In summary, this study evaluated the consistency of multiple LLMs with human raters in writing assessments, considering both scores and rationales. The results show that high-performing models such as GPT-4o and Claude~3.5 Sonnet not only scored accurately but also produced rationales that were semantically coherent and closely aligned with corresponding human rationales. The visualization of embedding reduction through PCA improves the interpretability of the scores assigned by both LLMs and human raters. These findings support the use of LLMs in automated scoring when both score consistency and reasoning similarity are strong.

This dual-level evaluation of both scoring accuracy and rationale similarity has implications for developing AI-supported assessment tools that prioritize transparency, interpretability, and fairness. As LLMs continue to influence automated scoring, models that provide not only a score but also a human-like justification for that score may begin to play a crucial role in supporting student learning and feedback.

Future work should be conducted to explore more diverse writing prompts, scoring rubrics, and languages, and to further examine the stability of rationale quality in the face of prompt revision and prompt engineering. Longitudinal studies could also be used to investigate how students respond to LLM-generated rationales and whether such feedback meaningfully impacts the development of students' writing skills.

\clearpage

\noindent \textbf{Funding Statement}\quad No funding was received for conducting this study.

\noindent \textbf{Data Availability}\quad The datasets generated and analyzed during the present study are not publicly available because of the conditions of the human-research ethical approvals.

\section*{Declarations}

\textbf{Competing Interests}\quad The authors declare no competing interests.

\begin{appendices}

\section{AI Training Protocol for Grading ER2 Student Samples}

\subsection{Prompt}

In this task, you will be asked to write in Chinese for a specific purpose and to a specific person. You should write in as complete and culturally appropriate a manner as possible, taking into account the purpose and the person described.

\textbf{Prompt:}

\begin{CJK*}{UTF8}{gbsn}
发件人：李晓红 \\
邮件主题：气候环境问题 \\

你好！我有个问题，想听听你的看法。今天上课时，老师讲到了我们这里的气候变化的问题，比如夏天变得越来越热，冬天的雪下得越来越少。老师还让我们写一篇作文，讨论气候的变化对我们生活的影响。你住的地方有哪些气候的变化？你认为这些变化对我们的生活有什么影响？谢谢！
\end{CJK*}

\subsection{Scoring Guidelines}

The scores are divided into six levels (0–6), and are evaluated based on the following three criteria:

\begin{CJK*}{UTF8}{gbsn}
\begin{itemize}
  \item \textbf{任务完成 (Task Completion)}
  \item \textbf{表达连贯性 (Delivery)}
  \item \textbf{语言使用 (Language Use)}
\end{itemize}
\end{CJK*}

---

\noindent\textbf{Score of 0: UNACCEPTABLE}  
\begin{itemize}
  \item Mere restatement of the prompt
  \item Clearly does not respond to the prompt; completely irrelevant to the topic
  \item Not in Mandarin Chinese
\end{itemize}

\textbf{Sample:}
\begin{quote}
\begin{CJK*}{UTF8}{gbsn}
我想 \\
\end{CJK*}
I don’t know
\end{quote}

---

\noindent\textbf{Score of 1: Very Weak}  
\textbf{Task Completion:} Minimal address of prompt. Disjointed or incoherent. \\
\textbf{Delivery:} Constant use of inappropriate register. \\
\textbf{Language Use:} Vocabulary/grammar errors that obscure meaning.

\textbf{Sample:}
\begin{quote}
\begin{CJK*}{UTF8}{gbsn}
嗨！对不起，我不知道你说什么。我的中文不好所以我不知道很多的汉字。\\
我觉气候变暖
\end{CJK*}
\end{quote}

---

\noindent\textbf{Score of 2: Weak}  
\textbf{Task Completion:} Only some aspects addressed. Scattered and fragmented ideas. \\
\textbf{Delivery:} Inappropriate register frequently. \\
\textbf{Language Use:} Frequent errors, limited vocabulary.

\textbf{Sample:}
\begin{quote}
\begin{CJK*}{UTF8}{gbsn}
好所以有地方有很长的夏天或者冬天。我住在一个地方夏天很长。\\
对不起！我很难理解你的电子邮件。我可能很多联系。
\end{CJK*}
\end{quote}

---

\noindent\textbf{Score of 3: Adequate}  
\textbf{Task Completion:} Directly addresses topic but may be incomplete. \\
\textbf{Delivery:} Inconsistently appropriate register. \\
\textbf{Language Use:} Simple structures; frequent but not meaning-blocking errors.

\textbf{Sample:}
\begin{quote}
\begin{CJK*}{UTF8}{gbsn}
我生活的地方，夏天很热，冬天越来越暖和。我觉得气候变化影响很多动物，也影响环境。
\end{CJK*}
\end{quote}

---

\noindent\textbf{Score of 4: Good}  
\textbf{Task Completion:} All aspects addressed, may lack elaboration. \\
\textbf{Delivery:} Generally coherent, loosely connected. \\
\textbf{Language Use:} Mostly appropriate with non-blocking errors.

\textbf{Sample:}
\begin{quote}
\begin{CJK*}{UTF8}{gbsn}
我住的地方气候变化也很明显。虽然夏天还是挺热的，但是我感觉冬天变得不太一样了……这些变化真的会影响我们的生活。
\end{CJK*}
\end{quote}

---

\noindent\textbf{Score of 5: Very Good}  
\textbf{Task Completion:} All aspects addressed clearly and logically. \\
\textbf{Delivery:} Connected discourse with occasional lapses. \\
\textbf{Language Use:} Appropriate vocabulary and grammar; minor errors.

\textbf{Sample:}
\begin{quote}
\begin{CJK*}{UTF8}{gbsn}
密歇根州有许多气候变化的例子。例如，冬天变得不那么寒冷……全球变暖可以影响我周围许多事物。
\end{CJK*}
\end{quote}

---

\noindent\textbf{Score of 6: Excellent}  
\textbf{Task Completion:} Fully addresses all aspects with detail. \\
\textbf{Delivery:} Logical, cohesive discourse. \\
\textbf{Language Use:} Rich vocabulary, minimal errors, wide range of structures.

\textbf{Sample:}
\begin{quote}
\begin{CJK*}{UTF8}{gbsn}
晓红你好，收到你的信让我很开心。刚好我的老师这个礼拜也在讲气候变化……\\
如果地球继续变暖，海平面上升，一些沿海城市甚至有可能被淹没！
\end{CJK*}
\end{quote}

\end{appendices}

\bibliography{bib}

\end{document}